%% file: main.tex
\def\year{2021}\relax
\documentclass[letterpaper]{article} % DO NOT CHANGE THIS
\usepackage{aaai21}  % DO NOT CHANGE THIS
\usepackage{times}  % DO NOT CHANGE THIS
\usepackage{helvet} % DO NOT CHANGE THIS
\usepackage{courier}  % DO NOT CHANGE THIS
\usepackage[hyphens]{url}  % DO NOT CHANGE THIS
\usepackage{graphicx} % DO NOT CHANGE THIS
\usepackage{natbib}  % DO NOT CHANGE THIS AND DO NOT ADD ANY OPTIONS TO IT
\usepackage{caption} % DO NOT CHANGE THIS AND DO NOT ADD ANY OPTIONS TO IT
\usepackage[switch]{lineno}
\usepackage{amsmath,amssymb,amsfonts}
\usepackage{algorithmic}
\usepackage{booktabs}
\usepackage{stmaryrd}
\usepackage{multirow}

\title{Layer-stacked Attention for Heterogeneous Network Embedding}
\author {
        Nhat Tran,\textsuperscript{\rm 1}
        Jean Gao \textsuperscript{\rm 1} \\
}
\affiliations {
	\textsuperscript{\rm 1} Dept. of Computer Science and Engineering, University of Texas at Arlington, USA \\
    \{nhat.tran, gao\}@uta.edu
}
\begin{document}
\maketitle

\begin{abstract}
The heterogeneous network is a robust data abstraction that can model entities of different types interacting in various ways. Such heterogeneity brings rich semantic information but presents nontrivial challenges in aggregating the heterogeneous relationships between objects -- especially those of higher-order indirect relations. Recent graph neural network approaches for representation learning on heterogeneous networks typically employ the attention mechanism, which is often only optimized for predictions based on direct links. Furthermore, even though most deep learning methods can aggregate higher-order information by building deeper models, such a scheme can diminish the degree of interpretability. To overcome these challenges, we explore an architecture--Layer-stacked ATTention Embedding (LATTE)--that automatically decomposes higher-order meta relations at each layer to extract the relevant heterogeneous neighborhood structures for each node. Additionally, by successively stacking layer representations, the learned node embedding offers a more interpretable aggregation scheme for nodes of different types at different neighborhood ranges. We conducted experiments on several benchmark heterogeneous network datasets. In both transductive and inductive node classification tasks, LATTE can achieve state-of-the-art performance compared to existing approaches, all while offering a lightweight model. With extensive experimental analyses and visualizations, the framework can demonstrate the ability to extract informative insights on heterogeneous networks.
\end{abstract}

\noindent Heterogeneous networks have been commonly used to model complex systems where there are multiple types of relationships among objects of different types. Such a rich semantic structure brings ripe graph mining opportunities for various real-world systems, including knowledge bases, academic networks, social networks, bioinformatic interactomes, and other multimodal abstractions. Recently, a significant line of research has been explored for representation learning of heterogeneous networks \cite{dong2020heterogeneous}. The basic principle behind these dimensionality-reduction approaches is to aggregate the high-dimensional information about a node's heterogeneous neighborhood to an embedding vector representation. These node embeddings can then aid in downstream machine learning tasks such as node classification, clustering, and link prediction.

Among the most effective approaches for representation learning on networks, graph neural network (GNN) methods has gained a dramatic increase in popularity in recent years
 \cite{kipf2016semi, hamilton2017inductive, velickovic2018graph}. While these powerful methods were designed for homogeneous networks, one can apply them to heterogeneous networks by ignoring the link/node type distinction and assuming the network structure to be homogeneous. However, this would be suboptimal, as it's been proven that neglecting the structural dependencies between relations by combining the multi-relations into a single network will omit important topological properties of the system \cite{battiston2014structural}. Therefore, the primary challenges for heterogeneous network embedding are maintaining the semantic information and aggregating the multi-relations for respective node types.

There have been several attempts to adopt GNNs to learn multi-relational networks \cite{schlichtkrull2018modeling, zhang2019heterogeneous}. More recently, several GNN models designed for heterogeneous networks have introduced the attention mechanism for increased interpretation of the aggregation of heterogeneous structures \cite{wang2019heterogeneous, yun2019graph, hu2020heterogeneous}.
However, these approaches for heterogeneous networks face at least one of the following issues.
First, some of them are only fitted to aggregate the multi-relations for a single primary node type; thus, they may require a manual design of meta paths.
Second, they only optimize for prediction between directly interacting nodes, which is insufficient to capture the heterogeneous network's global properties and higher-order structures. 
Third, although these GNNs with multiple hidden layers can flexibly propagate high-order information across layers, they do not explicitly preserve the semantics of higher-order meta relations.
These shortcomings can often affect the model's scalability, hinder the its generalizability for inductive predictions, or limit the interpretability of the model. 
% important to not only chooses relation-specific attention shared by all nodes, but also chooses attention differently for each node, to capture individual node heterogeneity

In consideration of these limitations and challenges, we aim to design an approach for heterogeneous GNNs to extract higher-order structures by leveraging the semantic information of all relations and node types. 
To handle heterogeneity in the network, we introduce a relation-specific attention mechanism, i.e., depending on the types and direction of a link. As each node type is involved in a subset of all relations, only the relevant relations are aggregated. The mechanism can then capture individual node heterogeneity, where each node is allowed to selectively determine which of its relation-specific neighborhoods contain a more prevalent signal.

To generate higher-order meta path connections between nodes of different types, we propose a novel scheme that combines transitive meta relations at each layer successively. As a result, all meta relation sequences of arbitrary length can be enumerated while retaining their semantic context. This process allows the model to distill the unique global structure of each node type by decomposing its heterogeneous neighborhoods at different ranges.
With a combination of the mechanisms proposed, our approach can generate higher-order meta paths and infer the most effective meta paths for prediction without requiring domain knowledge while maintaining the interpretability of the process.

Our main contributions with the proposed Layer-stacked ATTention Embedding (LATTE) method for heterogeneous networks are as followed. To the best of our knowledge, we are the first to introduce a GNN architecture that can successively stack hidden layers to generate higher-order meta relations specifically for each node type. Secondly, we utilize a aggregation technique that only combines specific meta relations depending on the node type. Lastly, we adopt a Noise Contrastive Estimation loss function to preserve the high-order proximities for weighted heterogeneous links, which improved performance for inductive node classification.

%\begin{itemize}
%	\item Propose an architecture that include both node-level and metapath-level attentions to effectively capture the heterogeneity among various node types and relation types in the network.
%	\item Through an efficient mechanism of stacking higher-order attention-based layers, LATTE can compute distant proximity between nodes connected through $n$-hop metapaths and can weigh the importance of various metapaths into consideration.
%	\item Formalizes a learning scheme that can simultaneously infer proximity-based pairwise link prediction and predict heterogeneous node representations for down-stream tasks.
%\end{itemize}

\section{Related Work}

\subsection{Graph Neural Networks}
In recent years, many classes of GNN methods have been developed for a variety of heterogeneous network types \cite{schlichtkrull2018modeling, zhang2019heterogeneous, wang2019heterogeneous, zhou2019hahe, hu2020heterogeneous}. Although these types of GNNs are flexible for end-to-end supervised prediction tasks, they only optimize for predictions between direct interactions. Compared to conventional network embedding methods \cite{grover2016node2vec, tang2015line}, standard GNNs generally do not take advantage of second-order relationships between indirect neighboring nodes. Recently, a paper by \cite{huang2020skipgnn} applied a fusion technique to combine first-order and second-order embeddings at alternating steps. Additionally, the Jumping Knowledge architecture from \cite{xu2018representation} and the GraphSAGE (sampling and aggregation) from \cite{hamilton2017inductive} has proposed to extend the neighborhood ranges; however, there has yet to be an extension of such techniques for heterogeneous networks.

Notably, GTN \cite{yun2019graph} was recently proposed to enable learning on higher-order meta paths in heterogeneous networks. It proposes a mechanism that soft-selects a convex combination of meta path layers using attention weights, then applies multiplication of adjacency matrices successively to reveal arbitrary-length transitive meta paths. This mechanism is unique in that it can infer attention weights not only on the given relations, but also on higher-order relations generated by deeper layers, a feature that most existing GNN methods often neglect. A few limitations with GTN is it necessarily assumes the feature distribution and representation space of different node and link types to be the same, and it cannot weigh the importance of each meta path separately for each node type. Additionally, GTN can be computationally expensive, since it requires computations involving the adjacency structure of all node types at once.

\subsection{Multiplex Network Embedding}
It is worth mentioning the approaches designed for a subclass of the heterogeneous network, the multiplex network. Many of the current multiplex or multiview network embedding methods \cite{fu2017hin2vec, matsuno2018mell, qu2017attention, shi2018mvn2vec} have proposed strategies for aggregating the learned embeddings of multiple network ``layers'' into a single unified embedding. This class of methods typically specify separate objectives for each of the layers to estimate the node features independently, then apply another objective to aggregate the information from all layers together.

Another paradigm is to use random-walk of meta paths to model heterogeneous structures, as proposed in \cite{perozzi2014deepwalk, dong2017metapath2vec, fu2017hin2vec}. This class of approaches can learn network representations without supervised training for a specific task. However, they only learn representations for the primary node type, which consequently requires the customized design of meta paths. Also, they can be sensitive to the random walk's hyper-parameter settings, which may introduce unwanted biases or is computationally costly, thus can lead to lacking performance. 
Another class of algorithm utilizing embedding translations can also be applied for embedding heterogeneous networks. For instance, \cite{bordes2013translating} learned linear transformations for each relation to model semantic relationships between entities. While embedding translations can effectively model heterogeneous networks, they are mainly fitted for link prediction tasks.

%Most existing multiplex network algorithms only aim to jointly embed multiple layers by considering the common patterns between layers. However, it’s possible that a pair of network layers can have non-complementary information, as they carry different semantic meanings. The mvn2vec method aims to address this problem by a “preservation” objective which captures the unique semantic meaning in each layer, and similarly, the MELL method introduces a layer-specific embedding. 

%It is important to closely consider two scenarios in a heterogeneous network: (1) a pair of layers have very similar interaction patterns while another pair of layers have disagreeing interaction patterns, and (2) for the same pair of layers, one portion of nodes have consistent interactions in different layers, while another portion of nodes have totally disagreeing interactions in different layers. 
%adopt MELL’s layer-specific embedding technique and combine with the attention mechanisms, which would allow each node to selectively determine which layer’s embedding information to be integrated.

\section{Method}
\subsection{Preliminary}
We consider a heterogeneous network as a complex system involving multiple types of links between nodes of various types. To effectively represent the complex structure of the system, it is important to define separate adjacency matrices to distinguish the nature of relationships. 
In this section, we define coherent notations to study the class of attributed heterogeneous networks.

%\begin{figure}[t]
%\centering
%\includegraphics[width=0.99\columnwidth]{figures/diagram.pdf}
%\caption{Conceptual illustration of the LATTE architecture demonstrating the layer-stacking operations that aggregates first-order and second-order meta relations. The heterogeneous network contains Paper-Author (PA), Paper-Conference (PC) and Paper-Term (PT) relations and their reverse relations (i.e. AP, CP, TP). The node feature inputs for each node types are $\mathbf{p}^0$, $\mathbf{a}^0$, $\mathbf{c}^0$, and $\mathbf{t}^0$, and the LATTE-$t$ embedding outputs for each respective node types are $\mathbf{p}^t$, $\mathbf{a}^t$, $\mathbf{c}^t$, and $\mathbf{t}^t$.}
%\label{concept}
%\end{figure}

\subsubsection{Definition 3.1: \textbf{Attributed Heterogeneous Network}}
is defined as a graph $G=(\mathcal{V}, \mathcal{E}, \mathcal{T})$ in which each node $i \in \mathcal{V}$ and each link $e_{ij} \in \mathcal{E}$ are associated with their mapping functions $\phi(i): \mathcal{V} \shortrightarrow \mathcal{T_V}$ and $\phi(e_{ij}): \mathcal{E} \shortrightarrow \mathcal{T_E}$. $\mathcal{T_V}$ and $\mathcal{T_E}$ denote the sets of object and relation types, where $|\mathcal{T_V}|+|\mathcal{T_E}|>2$. 
In the case of attributed heterogeneous network, the node features representation is given by $\Phi(i)= \mathbf{x_i} \in \mathbb{R}^{D_m}$, which maps node $i$ of node type $m \in \mathcal{T_V}$ to its corresponding feature vector $\mathbf{x_i}$ of dimension $D_m$.

We represent the heterogeneous link types as a set of biadjacency matrices $\mathcal{A} = \{ \mathbf{A}^{(m,n)} \mid \exists m, n \in \mathcal{T_V} \}$ where $|\mathcal{A}| = |\mathcal{T_E}|$. Each meta relation $(m,n)$ specifies a link type between source node type $m$ and target node type $n$, such that $\mathbf{A}^{(m,n)} = \{a^{(m,n)}_{ij} \mid i \in \mathcal{V}_m, j \in \mathcal{V}_n \}$. 
The biadjacency matrix may consist of weighted links, where $a^{(m,n)}_{ij} > 0$ if there exists a link, otherwise, $a^{(m,n)}_{ij}=0$, indicating the absence of evidence for interaction.
For a $\mathbf{A}^{(m,n)}$ subnetwork, we define node $i$'s neighbors set as $\mathcal{N}^{(m,n)}_{i}=\{j \mid \forall j \in \mathcal{V}_n \mathrel{s.t.} a^{(m,n)}_{ij}>0 \}$. Note that $\mathbf{A}^{(m,n)} \in \mathbb{R}^{|\mathcal{V}_m | \times |\mathcal{V}_n|}$'s size is non-quadratic, and thus does not have a diagonal. Furthermore, this definition assumes relations of directed links, but for a relation $\mathbf{A}^{(m,n)}$ with inherently undirected links, we may inject a reverse relation $\mathbf{A}^{(n, m)} = \{ a_{ji} | \forall a_{ij} \in \mathbf{A}^{(m,n)} \}$ into the $\mathcal{A}$ set.

\subsection{LATTE-1: First-order Heterogeneous Network Embedding}
In this section, we start by describing the attention-based layers used in the LATTE heterogeneous network embedding architecture. The attention mechanism utilized in our method closely follows GAT \cite{velickovic2018graph} but is extended to infer higher-order link proximity scores for nodes and links of heterogeneous types. We also introduce the layer building blocks where each layer has the roles of inferring node embeddings from heterogeneous node content and preserving higher-order link proximities.

The input to our model is the set of heterogeneous adjacency matrices $\mathcal{A}$ and the heterogeneous node features $\mathcal{X}=\{\mathbf{X}_m | \exists m \in \mathcal{T_V}\}$, where  $\mathbf{X}_m = \{\mathbf{x}_i\in \mathbb{R}^{D_m} | \forall i \in \mathcal{V}_m \}$. 
 At each $t^\text{th}$ layer,  we define the node embeddings output $\mathbf{h}^{t} \in \mathbb{R}^{|\mathcal{V}| \times F}$, where $F$ is the embedding dimension, as:
$$
	\mathbf{h}^{t} = f_t(\mathbf{h}^{t-1}, \mathcal{A}^t)
$$
where $\mathbf{h}_i^0 = \mathbf{x}_i$, and $\mathcal{A}^t$ is the heterogeneous link adjacency matrices in the $t^\text{th}$-order. In the next section, we describe the operations involved when $t=1$.

\subsubsection{Heterogeneous First-order Proximities} The first-order proximity refers to direct links between any two nodes in the network among the heterogeneous relations in $\mathcal{A}$.
In order to model the different distribution of links in each relation type $\mathbf{A}^{(m,n)} \in \mathcal{A}$, we utilize a node-level attentional kernel depending on the type of the relation. Additionally, to sufficiently encode node features into higher-level features, each node type $m$ requires a separate linear transformation applied to every node in $\mathcal{V}_m$. Given any node $i$ of type $m$ and node $j$ of type $n$, the respective kernel parameter $\mathbf{q}^1_{(m,n)}$ is utilized to compute the scoring mechanism:
\begin{equation}
	e^1_{ij} = {\mathbf{q}^1_{(m,n)}}^\top [\mathbf{U}^1_m \mathbf{x}_i || \mathbf{V}^1_n \mathbf{x}_j]
\end{equation}
where $\cdot^\top$ is the transposition and $||$ is the concatenation operation. We utilize two weight matrices $\mathbf{U}^1_m \in \mathbb{R}^{F \times D_m}$ and $\mathbf{V}^1_n \in \mathbb{R}^{F \times D_n}$ to obtain the "source" context and the "target" context, respectively, for a pair of nodes depending on the node types and the direction of the link. Note that the attention-based proximity score $e^1_{ij}$ is asymmetric, hence capable of modeling directed relationships where $a_{ij} \neq a_{ji}$.

\subsubsection{Inferring Node-level Attention Coefficients}
Next, our goal is to infer the importance of each neighbor node in the neighborhood around node $i$ for a given relation. Similar to GAT, we compute masked attention on existing links, such that $e^1_{ij}$ is only computed for first-order neighbor nodes $j \in \mathcal{N}^{(m,n)}_i$. The attention coefficients are computed by softmax normalization of the scores across all $j$, as:
\begin{equation}
	\alpha^{(m,n)}_{ij} = \frac{exp(\tau^{(m,n)} * e^1_{ij})}{\sum_{j'\in \mathcal{N}^{(m,n)}_i } exp(\tau^{(m,n)} * e^1_{ij'})}
\end{equation}
where $\tau^{(m,n)}$ is a learnable ``temperature`` variable initialized at $1$ that have the role of ``sharpening'' the attention scores \cite{chorowski2015attention} across the links distribution in the $(m,n)$ relation. It is expected that $\tau^{(m,n)}>1$ when the particular link distribution is dense or noisy, thus, integrating this technique allows the attention mechanism to focus on fewer neighbors.
Once obtained, the normalized attention coefficients are used to compute the features distribution of a node's by a linear combination of its neighbors for each relation.

\subsubsection{Inferring Relation Weighing Coefficients}
Since a node type $m$ is assumed to be involved in multiple types of relations, we must aggregate the relation-specific representations for each node. Previous works \cite{wang2019heterogeneous, yun2019graph} have proposed to measure the importance of each relation type by a set of semantic-level attention coefficients shared by all nodes. 
Instead, our method chooses to assign the relation attention coefficients \textit{differently} for each node $i$, which enables the capacity to capture individual node heterogeneity in the network.
First, we denote $\mathcal{A}_{(m \shortrightarrow)} \subset \mathcal{A}$ as the subset of meta relations with source type $m$. Since the number of relations involved in each node type can be different, each node of type $m$ only needs to soft-select from the subset of relevant relations. We utilize a linear transformation directly on node features to predict a normalized coefficient vector of size $|\mathcal{A}_{(m \shortrightarrow)}|+1$ that soft-selects among the set of associated relations $\mathcal{A}_{(m \shortrightarrow)}$ or itself. This operation is computed by:
\begin{equation}
	\boldsymbol{\beta}^{1, i} = softmax(\mathbf{W}^{1}_{m} \mathbf{x}_i + \mathbf{b}^1_m)
\end{equation}
where $\boldsymbol{\beta}^{1,i} \in \mathbb{R}^{|\mathcal{A}_{(m \shortrightarrow)}|+1}$ is parameterized by the weights $\mathbf{W}^{1}_{m} \in \mathbb{R}^{(|\mathcal{A}_{(m \shortrightarrow)}|+1) \times D_m}$ and bias $\mathbf{b}^1_m$ for each node type $m \in \mathcal{T_V}$.
 Since $\boldsymbol{\beta}^{1,i}$ is softmax normalized, $\mathbf{\beta}^{1,i}_0+ \sum^{\mathcal{A}_{(m \shortrightarrow)}}_{(m,n)} \mathbf{\beta}^{1,i}_{(m,n)} = 1$, where $\mathbf{\beta}^{1,i}_0$ is the coefficient indexed for the ``self'' choice.
 
\subsubsection{Aggregating First-order Neighborhoods}
It is important to not only capture the local neighborhood of a node in a single relation but also aggregate the neighborhoods among multiple relations and integrate the node's own features representation.
First, we gather information obtained from each relation's local neighborhoods, then combine their relation-specific embeddings. We apply both the node-level and relation-level attention coefficients to a weighted-average aggregation scheme:
\begin{equation}
	\mathbf{h}_i^1 = \sigma \left( \mathbf{\beta}^{1,i}_0 \mathbf{U}^1_m \mathbf{x}_i + \sum_{(m,n)}^{\mathcal{A}_{(m \shortrightarrow)}} \mathbf{\beta}^{1,i}_{(m,n)} \sum_{j}^{\mathcal{N}^{(m,n)}_i} \alpha^{(m,n)}_{ij}  \mathbf{V}^1_n \mathbf{x}_j \right)
\end{equation}
where $\sigma$ is a nonlinear function such as ReLU, and $i$'s node type is $m$.
The first-order node embedding is computed as an aggregation of linear-transformed immediate neighbor nodes.
Next, we show that multiple LATTE layers can be stacked successively in a manner that allows the attention mechanism to capture higher-order relationships.

\subsection{LATTE-\textit{T}: Higher-order Heterogeneous Network Embedding} In this section, we describe the layer-stacking operations involved to extract higher-order proximities when $t \geq 2$. The t$^\text{th}$-order proximity applies to indirect $t$-length metapaths achieved by combining two matching meta relations. For instance, when $t=2$, we can connect a relation $\mathbf{A}^{(m,n)} \in \mathcal{A}^{t-1}$ with target type $n$ to another relation $\mathbf{A}^{(n, p)} \in \mathcal{A}$ with matching source type $n$. Then, computing the Adamic-Adar \cite{adamic2003friends} as:
\begin{equation}
\begin{split}
\mathbf{A}^{(m, n, p)} & = \mathbf{A}^{(m, n)} \mathbf{D}^{-1} \mathbf{A}^{(n,p)}	 \\
\mathbf{D}_{jj} & = \sum_{i \in \mathcal{V}_m} a^{(m,n)}_{ij} + \sum_{k \in \mathcal{V}_p} a^{(n,p)}_{jk} 
\end{split}
\end{equation}
yields $\mathbf{A}^{(m, n, p)}$ as the degree-normalized biadjacency matrix consisting of length-2 metapaths from $\mathcal{V}_m$ nodes to $\mathcal{V}_p$ nodes. We define the set of meta relations containing all length-$t$ metapaths in the network, as:
\begin{equation}
\mathcal{A}^t = \mathcal{A}^{t-1} \times \mathcal{A}
\end{equation}
where $\times$ behaves as a cartesian product that yields the Adamic-Adar only for matching pairs of relations. A length-$t$ sequence of meta relations with source type $m$ and target type $p$ is denoted as $(m \overset{_t}\shortrightarrow p)$. This is directly applicable to the classical metapath paradigm \cite{sun2011pathsim}, where all possible $t$-length metapaths are decomposed in each separate relation in $\mathcal{A}^t$. Note that throughout this paper, the meta relations $(m,n)$ notation is overloaded for brevity. In fact, this architecture can handle multiple meta relation types with the same source type and target type, i.e. $\phi(e_{ij})= \left< \phi(i), \phi(e), \phi(j) \right>$, without loss of generalization.

\subsubsection{Heterogeneous Higher-order Proximities}
Learning the higher-order attention structure for $t^\text{th}$-order relations involves the composition between $\mathcal{A}^{t-1}$ and $\mathcal{A}$ meta relation sets. Since the t$^\text{th}$-order proximity is a measure between a node's $(t\text{-}1)^\text{th}$-order context to another node in the network, naturally, we must take into consideration of $\mathbf{h}^{t-1}$ as the prior-order context embeddings.  Similar to the first-order attention score, $e^t_{ik}$ is the $t^\text{th}$-order attention score between node $i \in \mathcal{V}_m$ and node $k \in \mathcal{V}_p$, defined as:
\begin{equation}
	e^t_{ik} = {\mathbf{q}^t_{(m \overset{_t}\shortrightarrow p)}}^\top [\mathbf{U}^t_m \mathbf{h}^{t-1}_i || \mathbf{V}^t_p \mathbf{x}_k]
\end{equation}
The $t^\text{th}$-order attention scoring mechanism is parameterized by 
$\mathbf{U}^t_m \in \mathbb{R}^{F \times F}$ and
$\mathbf{V}^t_p \in \mathbb{R}^{F \times D_p}$ for all node types, as well as 
$\mathbf{q}^t_{(m \overset{_t}\shortrightarrow p)} \in \mathbb{R}^{^{2F}}$ for each relation type in $\mathcal{A}^{t}$.
Then, in the same manner as in Eq. (3), the attention coefficients for each $t^\text{th}$-order neighbors in the relation is the softmax normalized $e^t_{ik}$ along with the temperature $\tau^{(m \overset{_t}\shortrightarrow p)}$:
$$
	\alpha^{(m \overset{_t}\shortrightarrow p)}_{ik} = \frac{exp(\tau^{(m \overset{_t}\shortrightarrow p)} * e^t_{ik})}{\sum_{k'\in \mathcal{N}^{(m \overset{_t}\shortrightarrow p)}_i } exp(\tau^{(m \overset{_t}\shortrightarrow p)} * e^t_{ik'})}
$$

Obtaining the relation-weighing coefficients in the $t^\text{th}$-order also involves the prior-order context embedding for each node.
For a node $i$ of type $m$, we apply the relation weighing mechanism using its prior-order embedding $\mathbf{h}_i^{t-1}$ with:
\begin{equation}
	\boldsymbol{\beta}^{t,i} = softmax(\mathbf{W}^{t}_{m} \mathbf{h}^{t-1}_i + \mathbf{b}^t_m)
\end{equation}
where $\boldsymbol{\beta}^{t,i} \in \mathbb{R}^{|\mathcal{A}^t_{(m \shortrightarrow)}|+1}$ is parameterized by weights $\mathbf{W}^{t}_{m} \in \mathbb{R}^{1+|\mathcal{A}^t_{(m \shortrightarrow)}| \times F}$. By far, LATTE can automatically identify important meta relations of any arbitrary $t$-length by learning an adaptive relation weighing mechanism.

\subsubsection{Aggregating Layer-wise Embeddings}
While the first-order embedding represents the local neighborhood among the multiple relations, its $t^\text{th}$-order embedding expands the receptive field's vicinity by traversing higher-order meta paths. The $t^\text{th}$-order embedding of node $i$ is expressed as:

\begin{table*}[ht!]
\centering
\resizebox{.95\linewidth}{!}{
\begin{tabular}{lllllllll}
\hline
\textbf{Dataset} & \textbf{\textit{Relations (A-B)}} & \textbf{\textit{\# nodes (A)}} & \textbf{\textit{\# nodes (B)}} & \textbf{\textit{\# links}} & \textbf{\textit{\# features}} & \textbf{\textit{Training}} & \textbf{\textit{Testing}} \\
\hline
\multirow{3}{*}{DBLP} & Paper-Author (PA) & 14328 & 4057 & 19645 & \multirow{3}{*}{334} & \multirow{3}{*}{20\%} & \multirow{3}{*}{70\%} \\
 & Paper-Conference (PC) & 14328 & 20 & 14328 & & & \\
 & Paper-Term (PT) & 14328 & 4057 & 88420 & & & \\
\hline
\multirow{2}{*}{ACM} & Paper-Author (PA) & 2464 & 5835 & 9744 & \multirow{2}{*}{1830} & \multirow{2}{*}{20\%} & \multirow{2}{*}{70\%} \\
 & Paper-Subject (PS) & 3025 & 56 & 3025 & & & \\
\hline
\multirow{2}{*}{IMDB} & Movie-Actor (MA) & 4780 & 5841 & 9744 & \multirow{2}{*}{1232} & \multirow{2}{*}{10\%} & \multirow{2}{*}{80\%} \\
 & Movie-Director (MD) & 4780 & 2269 & 3025 & & & \\
\hline
\end{tabular}}
\caption{Statistics for the heterogeneous network datasets.}
\label{degree_distribution}
\end{table*}

\begin{equation}
	\mathbf{h}_i^t = \sigma \left( \mathbf{\beta}^{t,i}_0 \mathbf{U}^t_m \mathbf{h}^{t-1}_i +
	\sum_{(m\overset{_t}\shortrightarrow p)}^{\mathcal{A}^t_{(m\shortrightarrow)}} \beta_{(m \overset{_t}\shortrightarrow p)}^{t,i} 
	\sum_{k}^{\mathcal{N}^{(m\overset{_t}\shortrightarrow p)}_i} \alpha^{(m\overset{_t}\shortrightarrow p)}_{ik} \mathbf{V}^t_p \mathbf{x}_k \right)
\end{equation}
With this framework, the receptive field of $t^\text{th}$-order relations is contained within each $t^\text{th}$-order context embedding. Furthermore, as $\boldsymbol{\beta}^{t,i}$ encapsulates each relation in $\mathcal{A}^t$ separately, it is possible to identify the specific relation types that are involved the composite representation. 

Given the layer-wise representations $\mathbf{h}^1_i,...,\mathbf{h}^T_i$ of node $i$, we obtain the final embedding output by concatenating all the $t$-order context embeddings, as:
\begin{equation}
	\mathbf{h}_i = \bigparallel_{t=1}^T f_t(\mathbf{h}^{t-1}, \mathcal{A}^t)  = \bigparallel_{t=1}^T \mathbf{h}^t_i
\end{equation}
where $\mathbf{h}_i \in \mathbb{R}^{TF}, \forall i \in \mathcal{V}$ with $T*F$ as the unified embedding dimension size for all node types.

\subsection{Preserving Proximities with Attention Scores}
We repurpose the computed attention scores to estimate the heterogeneous pairwise proximities in the network explicitly. Incorporating this objective not only enables our model for unsupervised learning but also allows the node-level attention mechanism to reinforce highly connected node pairs by taking advantage of weighted links. To preserve pairwise t\textsuperscript{th}-order proximities for all links in each $(m \overset{_t}\shortrightarrow p)$ relation, we apply the Noise Contrastive Estimation with negative sampling \cite{mikolov2013distributed} objective as:
\begin{equation}
\begin{split}
	L_t(\mathbf{A}^{(m \overset{_t}\shortrightarrow p)}) = & -\frac{1}{|\mathbf{A}^{(m \overset{_t}\shortrightarrow p)}|} \sum_{a_{ik}}^{\mathbf{A}^{(m \overset{_t}\shortrightarrow p)}} a_{ik} \log( \rho(e^t_{ik}) ) \\
	& - \frac{1}{K} \sum_k^K E_{a_{uv} \sim P(\mathbf{A}^{(m \overset{_t}\shortrightarrow p)})} [\log \rho(-e^t_{uv})] \\
\end{split}
\end{equation}
where $\rho$ denotes the sigmoid function applied to the attention score to infer a probability value. The first term models the observed links, the second term models the negative links drawn from the noise distribution in $(m \overset{_t}\shortrightarrow p)$, and $K$ is the number of sampled negative links. Typically, $K$ is chosen to be between 2 to 5 times the number of positive links.

\subsection{Model Optimization}
To learn from both the heterogeneous network's attributes and topology, we optimize the proximity-preserving objectives and the downstream objective of the embedding outputs with the standard back-propagation algorithm. For semi-supervised node classification, a multi-layer perceptron $g(\mathbf{h}_i) = \widetilde{\mathbf{y}}_i \in [0,1]^{G}$ follows the LATTE layers in order to predicts $G$ labels given the node embedding. The cross-entropy minimization objectives are defined as: 
\begin{equation}
	L(\mathcal{X},\mathcal{A}) = -\sum_{i \in \mathcal{V}_Y} \mathbf{y}_i log( g(\mathbf{h}_i) ) 
	+ \sum_{t=1}^T \sum^{\mathcal{A}^t}_{\mathbf{A}^{(m \overset{_t}\shortrightarrow n)}}L_t(\mathbf{A}^{(m \overset{_t}\shortrightarrow n)})
\end{equation}
where $\mathcal{V}_Y$ is the set of nodes that have labels, and $\mathbf{y}_i$ is the true label. 
The first term aims to encode the node embedding representations with attention mechanisms, while the second term reinforces the attention scores by iterating through weighted positive and sampled negative links. 

Our model allows for computing embeddings for a subnetwork each iteration; thus, it does not require computations involving the global network structure of all nodes at once. This approach not only enables mini-batch training on large networks that do not fit on memory but also makes our technique fitted for inductive learning. To perform online training at each iteration, an input batch is generated by recursively sampling a fixed number of neighbor nodes \cite{hamilton2017inductive}. Then, LATTE can yield embedding outputs for a sampled subnetwork given the local links and node attributes.

\section{Experiments}
An effective network representation learning method can generalize to an unseen node by accurately encoding its links and attributes and then ``aligning'' them to the embedding space learned from seen (trained) nodes.
In this section, we evaluate our method's effectiveness on several node classification experiments, where the task is to predict node labels for a portion of the network hidden during training.

\begin{table*}[ht]
\centering
\resizebox{.95\linewidth}{!}{
\begin{tabular}{lllllllll}
\hline
\textbf{Dataset} & \textbf{{Metric}} & \textbf{\textit{metapath2vec}} & \textbf{\textit{HIN2Vec}} & \textbf{\textit{HAN}} & \textbf{\textit{GTN}} & \textbf{\textit{LATTE-1}} & \textbf{\textit{LATTE-2}} & \textbf{\textit{LATTE-2$_{prox}$}}   \\
\hline
\multirow{3}{*}{DBLP} & \textit{F1$_{trans}$} & 0.7518 & 0.7431 & 0.9121 & 0.9203 & 0.8911$\pm$0.003 & \textbf{0.9240}$\pm$0.003 & 0.9156$\pm$0.003 \\
 & \textit{F1$_{induc}$} & -- & -- & 0.8666 & 0.8721 & 0.8620$\pm$0.004 & 0.8631$\pm$0.003 & \textbf{0.8822}$\pm$0.032 \\
 & \textit{\# params} & 2.3M & 2.3M & 240K & 125K & 78K & 111K & 111K \\
\hline
\multirow{3}{*}{ACM} & \textit{F1$_{trans}$} & 0.8879 & 0.8466 & 0.8725 & 0.9085 & 0.9118$\pm$0.005 & {0.9134}$\pm$0.005 & \textbf{0.9153}$\pm$0.003 \\
 & \textit{F1$_{induc}$} & -- & -- & 0.7909 & 0.8860 & 0.8988$\pm$0.003 & {0.9007}$\pm$0.003 & \textbf{0.9156}$\pm$0.003 \\
 & \textit{\# Params} & 387K & 1.1M & 1.5M & 326K & 250K & 273K & 273K \\
\hline
\multirow{3}{*}{IMDB} & \textit{F1$_{trans}$} & 0.4310 & 0.4404 & 0.5394 & 0.5924 & 0.6066$\pm$0.018 & {0.6135}$\pm$0.014 & \textbf{0.6363}$\pm$0.007 \\
 & \textit{F1$_{induc}$} & -- & -- & 0.3877 & 0.5810 & 0.6036$\pm$0.009 & {0.6117}$\pm$0.038 & \textbf{0.6355}$\pm$0.004 \\
 & \textit{\# Params} & 611K & 1.6M & 1.4M & 243K & 170K & 196K & 196K \\
\hline
\multicolumn{7}{l}{$^{\mathrm{\pm}}$ denotes the mean and standard deviation over 10 trials.}
\end{tabular}}
\caption{Performance comparison of Macro F1 for various methods over \textit{trans}-ductive and \textit{induc}-tive node classifications.}
\label{node_classification}
\end{table*}
\subsection{Datasets}
We conduct performance comparison experiments over several benchmark heterogeneous network datasets. In Table \ref{degree_distribution}, a summary of the network statistics is provided for each of the following datasets:
\begin{enumerate}
	\item \textbf{DBLP}\footnote{https://dblp.uni-trier.de}: a heterogenous network extracted from a bibliography dataset on major computer science journals and proceedings. The dataset have been preprocessed to contain 14328 \textit{papers}, 4057 \textit{authors}, 20 \textit{conferences}, and 8789 \textit{terms}. There are 3 relations types \textit{paper-author}, \textit{paper-conference} and \textit{paper-term} considered. The \textit{author}'s attributes are a bag-of-word representation of publication keywords. The classification task is to predict the label for each author among four domain areas: database, data mining, machine learning, and information retrieval.
	\item \textbf{ACM}\footnote{https://dl.acm.org}: A small citation network dataset containing \textit{paper-author} and \textit{paper-subject} relation types among 3025 \textit{papers}, 5835 \textit{authors}, and 56 \textit{subjects} node types. \textit{Paper} nodes are associated with a bag-of-words presentation of keywords as features. The task is to label the conference each paper is published in, among the KDD, SIGMOD, SIGCOM, MobiCOMM, and VLDB venues.
	\item \textbf{IMDB} \cite{cantador2011second}: A movie database network containing \textit{movie-actor} and \textit{movie-director} relations among 4780 \textit{movies}, 5841 \textit{actors}, and 2269 \textit{directors}. Each movie contain bag-of-words features of the plot, and the prediction task is to label the movie's genre among Action, Comedy, and Drama.
	\end{enumerate}
In each of the datasets, all directed relation have a reverse relation included. All self-loop links have been removed, unless if required for a certain algorithm.
	
\subsection{Experimental Setup}
To provide a consistent and reproducible experimental setup, the preprocessed networks were obtained from the CogDL Toolkit \cite{cogdl} benchmark datasets. Each of the datasets has been provided with a standard separation of train, validation, and test sets, as well as the full input features and labels set. Since our model evaluates these datasets based on their standard environment, the result from different experiments can be directly compared.

%\begin{figure*}[h!]
%\begin{tabular}{cc}
%  \includegraphics[width=70mm]{figures/blogcat.png} & \includegraphics[width=70mm]{figures/ogbn-mag.png} \\
%  
%  (a) BlogCatalog & (b) obgn-mag \\
%  \multicolumn{2}{c}{\includegraphics[width=\linewidth]{figures/ogbl-biokg.png}}\\
% \multicolumn{2}{c}{(c)} 
%\end{tabular}
%\caption{Degree distributions among different relation types}
%\label{degree_distribution}
%\end{figure*}

\subsubsection{Baselines}
We verify the effectiveness of our framework by testing multiple variants of LATTE along with other existing approaches.
For comparison with some of the state-of-the-art baselines, we consider various heterogeneous network embedding and GNN methods, including:
\begin{itemize}
	\item \textit{Metapath2Vec} \cite{dong2017metapath2vec}: An unsupervised random walk method that utilizes the skip-gram along with negative sampling on meta paths to embed heterogeneous nodes. It has been shown to achieve prominent performance among random walk based approaches.
	\item \textit{HIN2Vec} \cite{fu2017hin2vec}: a state-of-the-art deep neural network that learns embedding by considering the meta paths in an attributed heterogeneous network. It utilizes a random walk preprocessing, and it does not consider weighing of different meta paths.
	\item \textit{HAN} \cite{wang2019heterogeneous}: A GNN that employs a GAT-based node-level attention mechanism for heterogeneous networks. It proposes a hierarchical attention procedure that weighs the importance for each meta path, however only among pre-defined hand-crafted meta paths.
	\item \textit{GTN} \cite{yun2019graph}: A GNN with an attention mechanism that weighs and combines heterogeneous meta paths successively into higher-order structures, then performs graph convolution on the resulting adjacency matrix.
	\item \textit{LATTE-1}: A variant of the proposed LATTE model with one layer that only considers first-order meta relations. The pairwise proximity preserving objectives is excluded.
	\item \textit{LATTE-2}: A variant of LATTE with two layers that considers both first-order and second-order meta relations. The pairwise proximity preserving objectives is excluded.
	\item \textit{LATTE-2$_{prox}$}: Same as \textit{LATTE-2} but additionally optimizes the higher-order proximity preserving objectives.
\end{itemize}
Every method was evaluated on the identical split of training, validation, and testing sets for fairness and reproducibility. The final model is trained on the training set until the early stopping criteria on the validation set is met, then evaluated on the test set. Additionally, each method must exploit all relations and the available node attributes in the dataset, except for \textit{metapath2vec} due to its limitation. If a particular node type in the heterogeneous network is not attributed, we instantiate a set of learnable embeddings to replace $\mathcal{X}$ as node features.

\subsubsection{Implementation Details}
We set the following hyper-parameters identically for all methods: embedding dimension size at 128, learning rate at 0.001, mini-batch size at 2048, and early stopping if the validation loss doesn't decrease after ten epochs. For HAN and GTN, the number of GNN hidden layers is 2, preceding an MLP that predicts node labels given the embedding outputs in an end-to-end manner. For random walk based methods, a logistic classifier is employed to perform node classification given the learned node embeddings. The hyper-parameters for metapath2vec and HIN2Vec are walk length at 100, window size at 5, walks per node at 40, and the number of negative samples at 5. Among GNN-based methods, the batch sampling procedure that recursively samples a fixed number of neighbor nodes \cite{hamilton2017inductive} is utilized, with neighborhood sample sizes 25 and 20. Where possible, the standard implementation of baseline methods has been provided by the CogDL Toolkit.

For all LATTE variants, the best performing hyper-parameters selected ReLU as the embedding activation function, drop-out at 30\% on the embedding outputs, and weight decay regularization (excluding biases) at 0.01. In LATTE-2$_{prox}$, the negative sampling ratio is set to $5.0$. 
The models have been implemented with Pytorch Geometric (PyG), and the experiments have been conducted on a GeForce RTX 2080 Ti with 11 GB of GPU memory. The hyper-parameter tuning were conducted by Weight and Biases \cite{wandb}, and the parameter ranges tested were reported in the technical appendix. 

%We have tested for the multi-head attention technique \cite{vaswani2017attention} usually used in many attention-based methods; however, it only provided a negligible improvement in performance in our experiments.

\subsection{Node Classification Experiments and Results}
We consider the semi-supervised classification tasks in both inductive and transductive settings to perform thorough evaluations of representation learning in heterogeneous networks.
In the transductive setting, models can traverse on the subgraph containing nodes in the test set during training. In contrast, the inductive setting requires the models never to encounter the test subgraph during the training phase and must predict testing nodes' labels on the novel subgraph at the testing phase. We train and evaluate all baseline methods to predict test nodes for each transductive and inductive setting over ten trials.

To measure the classification performance of the prediction outputs, we record the precision and recall for each class label to compute the F1 score. Due to the apparent class imbalance in the three datasets, we report only the averaged Macro-F1 score, which was the more challenging metric in similar experiments \cite{wang2019heterogeneous}. The performance comparisons are reported in Table \ref{node_classification}. For {metapath2vec}, {HIN2Vec}, {HAN}, and {GTN}, the benchmark Macro F1 scores in the transductive setting has been provided by the CogDL Toolkit, while the Macro F1 in the inductive setting are averaged scores over 10 experiment runs.

The top performance by {LATTE}-2$_{prox}$ indicates its effectiveness at learning node representations on the high-order meta relation structures, especially with 80-90\% of the network set aside for testing. Compared to HAN, which does not consider higher-order relations, GTN and LATTE-2 have a significant edge in inductive prediction because both can capture global properties. Compared to GTN, which does not maintain the semantic space of individual meta path, {LATTE}-2$_{prox}$ outperforms with explicit proximity-preserving objectives for each of the decomposed higher-order meta relations.

\subsection{Interpretation of the Attention Mechanism}
\begin{figure}[t!]
\centering
\includegraphics[width=0.64\columnwidth]{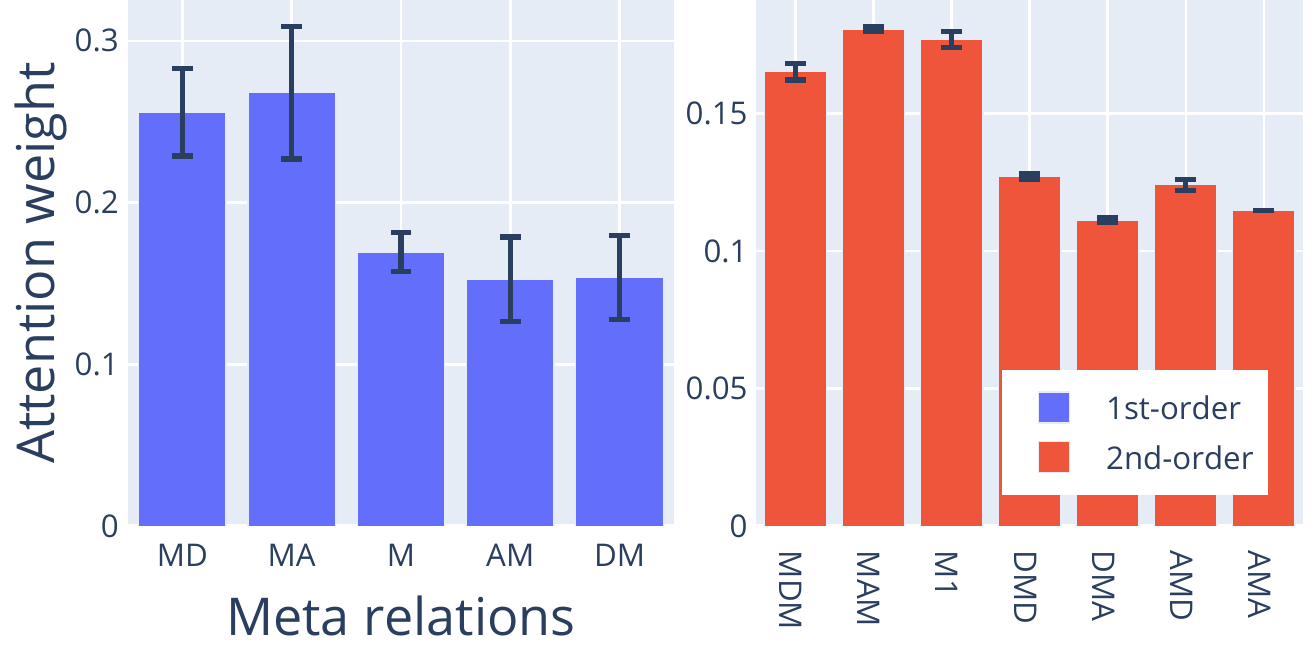} \includegraphics[width=0.34\columnwidth]{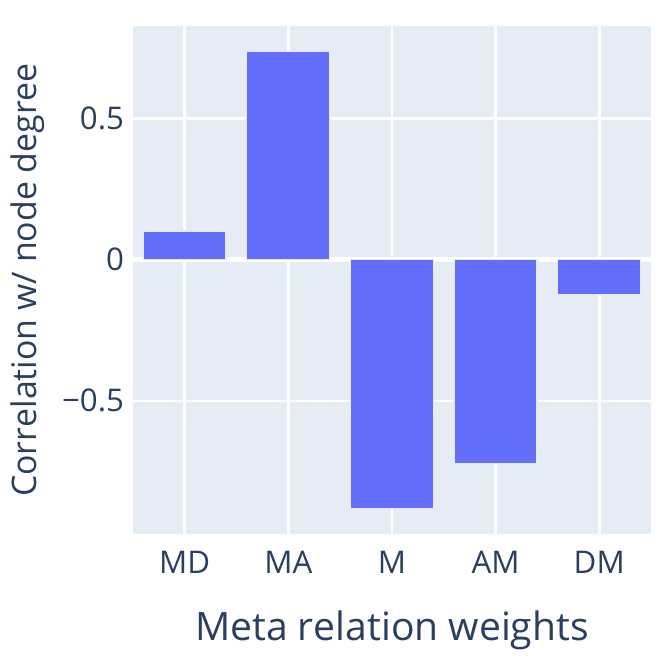}
\begin{tabular}{ccccccccccccccc}
&&&& (a) &&&&&&&&& (b) \\[3pt]
\end{tabular}
\caption{(a) Average and standard deviation of the 1st and 2nd-order meta relation attention weights, where relations starting with \textit{M} are aggregated to embed IMDB movie nodes. (b) Correlation between nodes degrees and relation weights for each meta relation in IMDB. A single-letter relation (e.g. \textit{M}, \textit{M1}) denotes the ``self'' choice. }
\label{deg_corr}
\end{figure}

LATTE's fundamental properties are the construction of higher-order meta relations and the attention mechanism that weighs the importance of those relations. To demonstrate these features' benefits, we interpret the importance levels chosen for each meta relations and verify whether they reflect the structural topology in the heterogeneous network. Given the learned weights $\boldsymbol{\beta}^{t,i}$ for each node $i$ at a layer $t$, we can assess not only the averaged meta relation weights for a node type, but also the individual meta relation weights for each node. In Fig. \ref{deg_corr}, we report the average and standard deviation of the meta relation attention weights for IMDB, as well as the correlation between those weights and the node degrees for each relation. The meta relation weights for DLBP and ACM are reported as supplementary material.

For IMDB movies, it can be observed that on average, the \textit{MA}, \textit{MD}, \textit{MDM}, and \textit{MAM} meta relations have the highest attention weights. This indicates that information from the \textit{movie-actor} neighborhoods, \textit{movie-director} neighborhoods, and node's features are relatively more represented in each \textit{movie}'s first-order embedding. This selection also persists in the second-order embeddings, where \textit{MDM} and \textit{MAM} have higher weights. Additionally, when looking at the correlation between \textit{MA}'s weights and the degree of \textit{MA} links over all nodes, there is a $0.73$ correlation, which indicates the attention mechanism can adaptively weigh the relation based on the number connections present in the node. Interestingly, there is a substantial negative correlation of $-0.88$ between the \textit{M} ``self'' relation weights and the node degree. This fact indicates that nodes with fewer or no links will choose a higher weight for its own features, since little information can be gained from other modalities. As individual nodes may have varying levels of participation among the various relations, this result demonstrates that LATTE can select the most effective meta relation for individual nodes depending on its local and global properties in the heterogeneous topology.

\section{Discussion and Conclusion}
The task of aggregating heterogeneous relations remains a fundamental challenge in designing a representation learning method for heterogeneous networks. Multiple relations can represent different semantics, and their link distributions can be overlapping, interconnected, or non-complementary. Therefore, it is an appropriate first step to consider them as separate components of the network to unravel their structural dependencies. One of the key differences between existing GNN methods and the proposed LATTE is that the latter exploits the semantic information in each meta relation. Instead of conflating heterogeneous relations for all node types as in HAN and GTN, LATTE aggregates only the relevant relations for each node type. Furthermore, by considering the source type and target type of each meta relation, only relevant pairs of relations can be joined during generating higher-order meta paths. A significant benefit to this approach is that it relieves the computational burden of multiplying adjacency matrices for all nodes while allowing distinct representation for the different node types.
 
This work has proposed an architecture for heterogeneous network embedding, which can generate higher-order meta relations. The benefits of the mechanism proposed are not only to improve inductive node classification performance but also to improve interpretation of deep GNN models.
In the future, we will explore whether to incorporate a self-attention mechanism to learn the structural dependencies between relations by  propagating information between the different relation-specific embeddings. Other interesting future developments are to enable LATTE to pre-train without supervision and to extend LATTE to link prediction tasks.

%Another interesting future development is a network sampling procedure that can efficiently extract higher-order relations, such as a random walk-based approach. By feeding the higher-order links directly to each $t^{th}$-layer, it would mitigate the computational burden of multiplying successive adjacency matrices.
%Another future direction is to incorporate a self-attention mechanism, where information can flow between the different relation-specific embeddings and learn structural dependencies between relations. 

%\section{Acknowledgments}
\bigskip

\bibliography{cites}
\include{technical_appendix}

\end{document}

%% file: technical_appendix.tex
\onecolumn
\section{Technical Appendix}

\begin{figure}[h]
\centering
\includegraphics[width=0.4\linewidth]{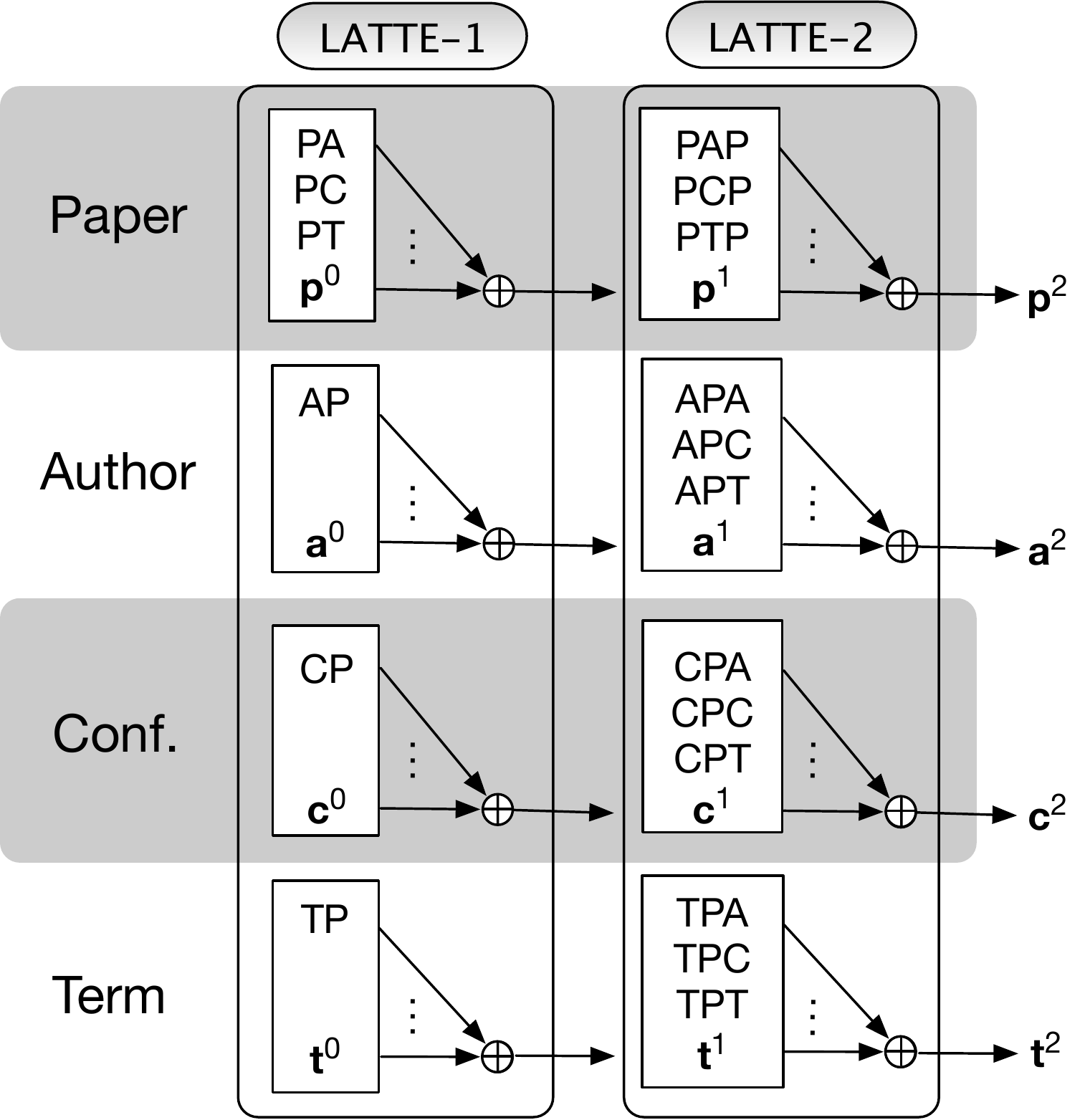}
\caption{Conceptual illustration of the LATTE architecture demonstrating the layer-stacking operations that aggregates first-order and second-order meta relations. The heterogeneous network contains Paper-Author (PA), Paper-Conference (PC) and Paper-Term (PT) relations, in addition to their reverse relations (i.e. AP, CP, TP). The node feature inputs for each node types are $\mathbf{p}^0$, $\mathbf{a}^0$, $\mathbf{c}^0$, and $\mathbf{t}^0$, and the LATTE-$t$ embedding outputs for each respective node types are $\mathbf{p}^t$, $\mathbf{a}^t$, $\mathbf{c}^t$, and $\mathbf{t}^t$. The t$^{th}$-order meta relations are generated by combining relations from $\mathcal{A}^{t-1}$ and $\mathcal{A}$.}
\label{concept}
\end{figure}

\subsection{Hyper-parameter Tuning}
The hyper-parameter tuning were conducted by the Weight and Biases\footnote{Biewald, L. 2020. Experiment Tracking with Weights and Biases. URL https://www.wandb.com/. Software available from wandb.com.} platform, where we utilize a random search approach that chooses random sets of parameter values. The parameters tested for are the embedding dimension, ${t}$-order, attention scores activation function, number of neighbors sampled, negative sampling ratio, embedding output activation function, and dropout probability. 

\begin{figure}[h]
\centering
\includegraphics[width=0.9\linewidth]{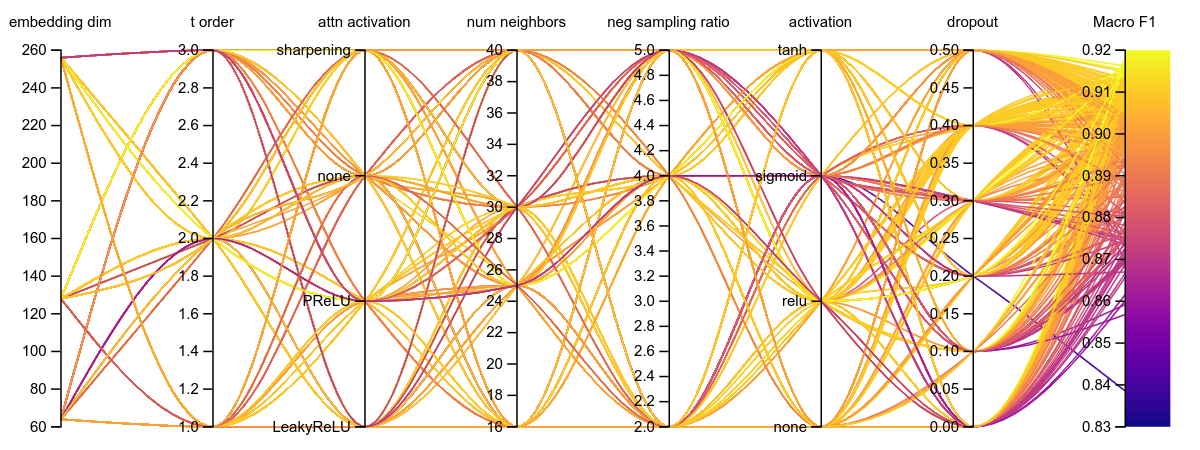}
\caption{Hyper-parameters tuning for Macro F1 performance on ACM (inductive) dataset. The lighter colors indicate trial runs which has a higher Macro F1 score.}
\label{hparams}
\end{figure}

\subsection{Interpretation of the Attention Mechanism for DBLP and ACM}
\begin{figure}[ht]
\centering
\begin{tabular}{ccc}
\includegraphics[width=0.6\linewidth]{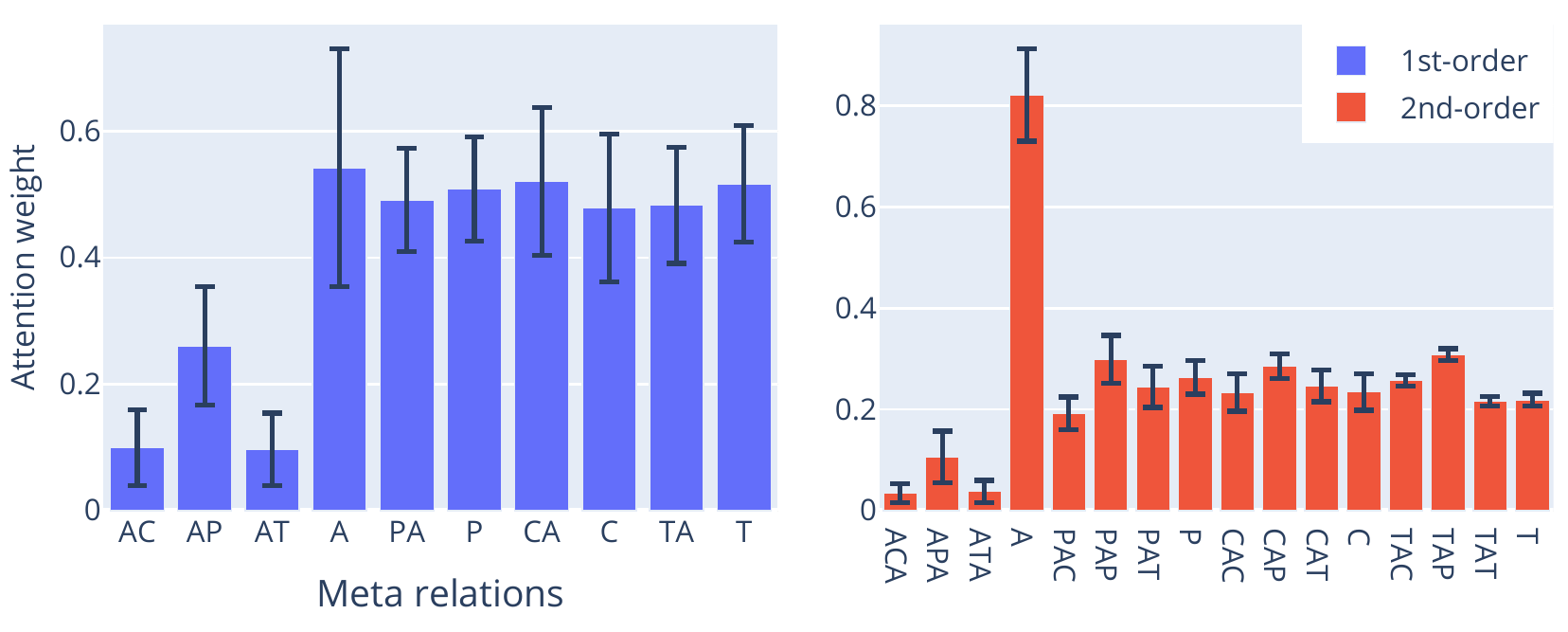} && \includegraphics[width=0.24\linewidth]{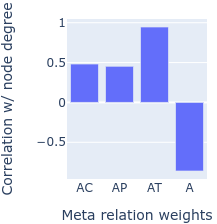} \\
(a) && (b) \\
\end{tabular}
\caption{(a) Average and standard deviation of the 1st and 2nd-order meta relation attention weights for DBLP dataset. \quad\quad\quad (b) Correlation between nodes degrees and relation weights for each meta relation in DBLP.}
\label{dblp}
\end{figure}

\begin{figure}[ht]
\centering
\begin{tabular}{ccc}
\includegraphics[width=0.6\linewidth]{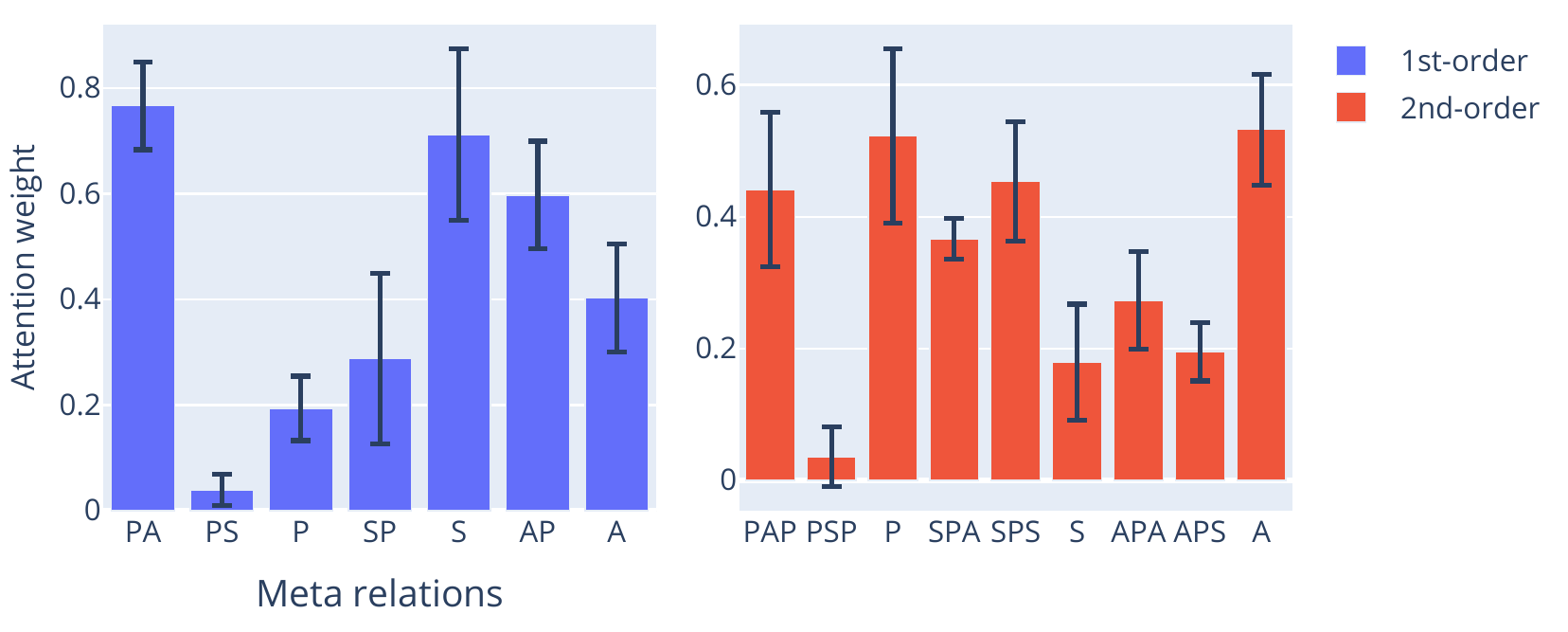} && \includegraphics[width=0.24\linewidth]{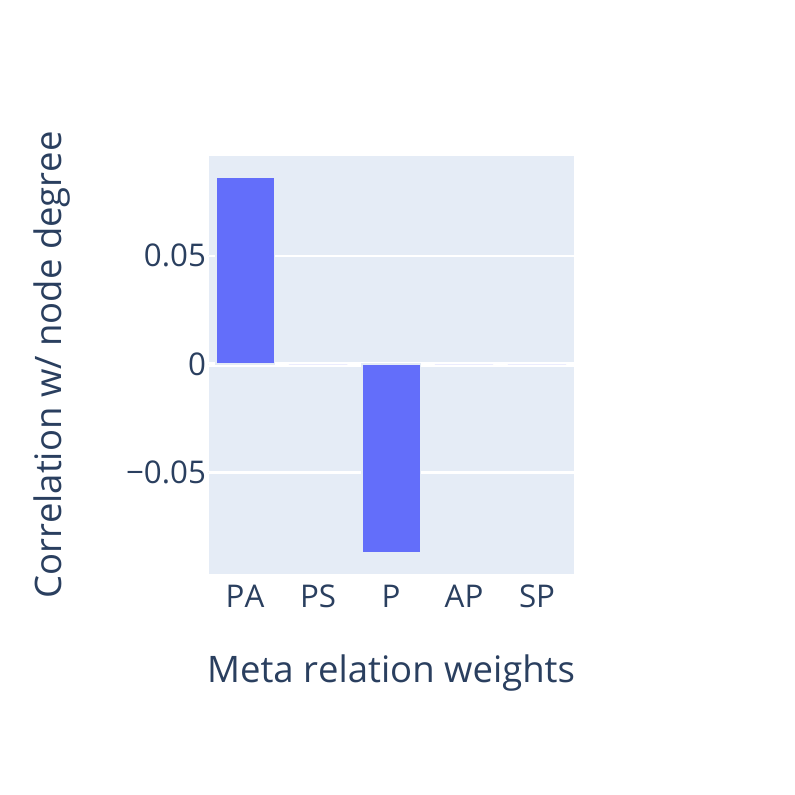} \\
(a) && (b) \\
\end{tabular}
\caption{(a) Average and standard deviation of the 1st and 2nd-order meta relation attention weights for ACM dataset. \quad\quad\quad (b) Correlation between nodes degrees and relation weights for each meta relation in ACM.}
\label{acm}
\end{figure}

Following the demonstration to interpret the learned attention weights in the IMDB dataset, we report the same attention weights and the weights-degree correlation results for the DBLP and ACM datasets. In Fig. \ref{dblp} and \ref{acm}, it can be observed that correlation between the meta relation weights and the node degree exhibits the same phenomenon described for IMDB.

\bigskip

%% file: main.bbl
\begin{thebibliography}{29}
\providecommand{\natexlab}[1]{#1}
\providecommand{\url}[1]{\texttt{#1}}
\providecommand{\urlprefix}{URL }
\expandafter\ifx\csname urlstyle\endcsname\relax
  \providecommand{\doi}[1]{doi:\discretionary{}{}{}#1}\else
  \providecommand{\doi}{doi:\discretionary{}{}{}\begingroup
  \urlstyle{rm}\Url}\fi

\bibitem[{Adamic and Adar(2003)}]{adamic2003friends}
Adamic, L.~A.; and Adar, E. 2003.
\newblock Friends and neighbors on the web.
\newblock \emph{Social networks} 25(3): 211--230.

\bibitem[{Battiston, Nicosia, and Latora(2014)}]{battiston2014structural}
Battiston, F.; Nicosia, V.; and Latora, V. 2014.
\newblock Structural measures for multiplex networks.
\newblock \emph{Physical Review E} 89(3): 032804.

\bibitem[{Biewald(2020)}]{wandb}
Biewald, L. 2020.
\newblock Experiment Tracking with Weights and Biases.
\newblock \urlprefix\url{https://www.wandb.com/}.
\newblock Software available from wandb.com.

\bibitem[{Bordes et~al.(2013)Bordes, Usunier, Garcia-Duran, Weston, and
  Yakhnenko}]{bordes2013translating}
Bordes, A.; Usunier, N.; Garcia-Duran, A.; Weston, J.; and Yakhnenko, O. 2013.
\newblock Translating embeddings for modeling multi-relational data.
\newblock In \emph{Advances in neural information processing systems},
  2787--2795.

\bibitem[{Cantador, Brusilovsky, and Kuflik(2011)}]{cantador2011second}
Cantador, I.; Brusilovsky, P.; and Kuflik, T. 2011.
\newblock Second workshop on information heterogeneity and fusion in
  recommender systems (HetRec2011).
\newblock In \emph{Proceedings of the fifth ACM conference on Recommender
  systems}, 387--388.

\bibitem[{Cen et~al.(2020)Cen, Wang, Hou, Chen, and Tang}]{cogdl}
Cen, Y.; Wang, Y.; Hou, Z.; Chen, Q.; and Tang, J. 2020.
\newblock CogDL: An Extensive Research Toolkit for Deep Learning on Graphs.
\newblock \urlprefix\url{https://github.com/thudm/cogdl}.

\bibitem[{Chorowski et~al.(2015)Chorowski, Bahdanau, Serdyuk, Cho, and
  Bengio}]{chorowski2015attention}
Chorowski, J.~K.; Bahdanau, D.; Serdyuk, D.; Cho, K.; and Bengio, Y. 2015.
\newblock Attention-based models for speech recognition.
\newblock In \emph{Advances in neural information processing systems},
  577--585.

\bibitem[{Dong, Chawla, and Swami(2017)}]{dong2017metapath2vec}
Dong, Y.; Chawla, N.~V.; and Swami, A. 2017.
\newblock metapath2vec: Scalable representation learning for heterogeneous
  networks.
\newblock In \emph{Proceedings of the 23rd ACM SIGKDD international conference
  on knowledge discovery and data mining}, 135--144.

\bibitem[{Dong et~al.(2020)Dong, Hu, Wang, Sun, and
  Tang}]{dong2020heterogeneous}
Dong, Y.; Hu, Z.; Wang, K.; Sun, Y.; and Tang, J. 2020.
\newblock Heterogeneous Network Representation Learning.
\newblock IJCAI.

\bibitem[{Fu, Lee, and Lei(2017)}]{fu2017hin2vec}
Fu, T.-y.; Lee, W.-C.; and Lei, Z. 2017.
\newblock Hin2vec: Explore meta-paths in heterogeneous information networks for
  representation learning.
\newblock In \emph{Proceedings of the 2017 ACM on Conference on Information and
  Knowledge Management}, 1797--1806.

\bibitem[{Grover and Leskovec(2016)}]{grover2016node2vec}
Grover, A.; and Leskovec, J. 2016.
\newblock node2vec: Scalable feature learning for networks.
\newblock In \emph{Proceedings of the 22nd ACM SIGKDD international conference
  on Knowledge discovery and data mining}, 855--864.

\bibitem[{Hamilton, Ying, and Leskovec(2017)}]{hamilton2017inductive}
Hamilton, W.; Ying, Z.; and Leskovec, J. 2017.
\newblock Inductive representation learning on large graphs.
\newblock In \emph{Advances in neural information processing systems},
  1024--1034.

\bibitem[{Hu et~al.(2020)Hu, Dong, Wang, and Sun}]{hu2020heterogeneous}
Hu, Z.; Dong, Y.; Wang, K.; and Sun, Y. 2020.
\newblock Heterogeneous graph transformer.
\newblock In \emph{Proceedings of The Web Conference 2020}, 2704--2710.

\bibitem[{Huang et~al.(2020)Huang, Xiao, Glass, Zitnik, and
  Sun}]{huang2020skipgnn}
Huang, K.; Xiao, C.; Glass, L.; Zitnik, M.; and Sun, J. 2020.
\newblock SkipGNN: Predicting Molecular Interactions with Skip-Graph Networks.
\newblock \emph{arXiv preprint arXiv:2004.14949} .

\bibitem[{Kipf and Welling(2016)}]{kipf2016semi}
Kipf, T.~N.; and Welling, M. 2016.
\newblock Semi-supervised classification with graph convolutional networks.
\newblock \emph{arXiv preprint arXiv:1609.02907} .

\bibitem[{Matsuno and Murata(2018)}]{matsuno2018mell}
Matsuno, R.; and Murata, T. 2018.
\newblock MELL: effective embedding method for multiplex networks.
\newblock In \emph{Companion Proceedings of the The Web Conference 2018},
  1261--1268.

\bibitem[{Mikolov et~al.(2013)Mikolov, Sutskever, Chen, Corrado, and
  Dean}]{mikolov2013distributed}
Mikolov, T.; Sutskever, I.; Chen, K.; Corrado, G.~S.; and Dean, J. 2013.
\newblock Distributed representations of words and phrases and their
  compositionality.
\newblock In \emph{Advances in neural information processing systems},
  3111--3119.

\bibitem[{Perozzi, Al-Rfou, and Skiena(2014)}]{perozzi2014deepwalk}
Perozzi, B.; Al-Rfou, R.; and Skiena, S. 2014.
\newblock Deepwalk: Online learning of social representations.
\newblock In \emph{Proceedings of the 20th ACM SIGKDD international conference
  on Knowledge discovery and data mining}, 701--710.

\bibitem[{Qu et~al.(2017)Qu, Tang, Shang, Ren, Zhang, and
  Han}]{qu2017attention}
Qu, M.; Tang, J.; Shang, J.; Ren, X.; Zhang, M.; and Han, J. 2017.
\newblock An attention-based collaboration framework for multi-view network
  representation learning.
\newblock In \emph{Proceedings of the 2017 ACM on Conference on Information and
  Knowledge Management}, 1767--1776.

\bibitem[{Schlichtkrull et~al.(2018)Schlichtkrull, Kipf, Bloem, Van Den~Berg,
  Titov, and Welling}]{schlichtkrull2018modeling}
Schlichtkrull, M.; Kipf, T.~N.; Bloem, P.; Van Den~Berg, R.; Titov, I.; and
  Welling, M. 2018.
\newblock Modeling relational data with graph convolutional networks.
\newblock In \emph{European Semantic Web Conference}, 593--607. Springer.

\bibitem[{Shi et~al.(2018)Shi, Han, He, He, Yang, Luo, and
  Han}]{shi2018mvn2vec}
Shi, Y.; Han, F.; He, X.; He, X.; Yang, C.; Luo, J.; and Han, J. 2018.
\newblock mvn2vec: Preservation and collaboration in multi-view network
  embedding.
\newblock \emph{arXiv preprint arXiv:1801.06597} .

\bibitem[{Sun et~al.(2011)Sun, Han, Yan, Yu, and Wu}]{sun2011pathsim}
Sun, Y.; Han, J.; Yan, X.; Yu, P.~S.; and Wu, T. 2011.
\newblock Pathsim: Meta path-based top-k similarity search in heterogeneous
  information networks.
\newblock \emph{Proceedings of the VLDB Endowment} 4(11): 992--1003.

\bibitem[{Tang et~al.(2015)Tang, Qu, Wang, Zhang, Yan, and Mei}]{tang2015line}
Tang, J.; Qu, M.; Wang, M.; Zhang, M.; Yan, J.; and Mei, Q. 2015.
\newblock Line: Large-scale information network embedding.
\newblock In \emph{Proceedings of the 24th international conference on world
  wide web}, 1067--1077.

\bibitem[{Veli{\v{c}}kovi{\'{c}} et~al.(2018)Veli{\v{c}}kovi{\'{c}}, Cucurull,
  Casanova, Romero, Li{\`{o}}, and Bengio}]{velickovic2018graph}
Veli{\v{c}}kovi{\'{c}}, P.; Cucurull, G.; Casanova, A.; Romero, A.; Li{\`{o}},
  P.; and Bengio, Y. 2018.
\newblock {Graph Attention Networks}.
\newblock \emph{International Conference on Learning Representations} .

\bibitem[{Wang et~al.(2019)Wang, Ji, Shi, Wang, Ye, Cui, and
  Yu}]{wang2019heterogeneous}
Wang, X.; Ji, H.; Shi, C.; Wang, B.; Ye, Y.; Cui, P.; and Yu, P.~S. 2019.
\newblock Heterogeneous graph attention network.
\newblock In \emph{The World Wide Web Conference}, 2022--2032.

\bibitem[{Xu et~al.(2018)Xu, Li, Tian, Sonobe, Kawarabayashi, and
  Jegelka}]{xu2018representation}
Xu, K.; Li, C.; Tian, Y.; Sonobe, T.; Kawarabayashi, K.-i.; and Jegelka, S.
  2018.
\newblock Representation learning on graphs with jumping knowledge networks.
\newblock \emph{arXiv preprint arXiv:1806.03536} .

\bibitem[{Yun et~al.(2019)Yun, Jeong, Kim, Kang, and Kim}]{yun2019graph}
Yun, S.; Jeong, M.; Kim, R.; Kang, J.; and Kim, H.~J. 2019.
\newblock Graph transformer networks.
\newblock In \emph{Advances in Neural Information Processing Systems},
  11983--11993.

\bibitem[{Zhang et~al.(2019)Zhang, Song, Huang, Swami, and
  Chawla}]{zhang2019heterogeneous}
Zhang, C.; Song, D.; Huang, C.; Swami, A.; and Chawla, N.~V. 2019.
\newblock Heterogeneous graph neural network.
\newblock In \emph{Proceedings of the 25th ACM SIGKDD International Conference
  on Knowledge Discovery \& Data Mining}, 793--803.

\bibitem[{Zhou et~al.(2019)Zhou, Bu, Wang, Chen, and Wang}]{zhou2019hahe}
Zhou, S.; Bu, J.; Wang, X.; Chen, J.; and Wang, C. 2019.
\newblock HAHE: Hierarchical attentive heterogeneous information network
  embedding.
\newblock \emph{arXiv preprint arXiv:1902.01475} .

\end{thebibliography}
